Classification of Dermoscopy Images using Deep Learning

ND Reddy, BS[1]

1. *Tufts University School of Medicine, Boston, MA*

**ABSTRACT**

Skin cancer is one of the most common forms of cancer and its incidence is projected to rise over the next decade. Artificial intelligence is a viable solution to the issue of providing quality care to patients in areas lacking access to trained dermatologists. Considerable progress has been made in the use of automated applications for accurate classification of skin lesions from digital images. In this manuscript, we discuss the design and implementation of a deep learning algorithm for classification of dermoscopy images from the HAM10000 Dataset. We trained a convolutional neural network based on the ResNet50 architecture to accurately classify dermoscopy images of skin lesions into one of seven disease categories. Using our custom model, we obtained a balanced accuracy of 91% on the validation dataset.

**INTRODUCTION**

Skin cancer is one of the most common forms of cancer. In the United States, non-melanoma skin cancer (NMSC) accounts for the majority of skin cancer diagnoses, approximately 5.4 million cases on a yearly basis[1]. Although melanoma comprises of only 1% of all skin cancer cases in the United States, it is associated with significantly higher mortality of over 9000 deaths each year[2]. The incidence rates for NMSC and melanoma have risen at an alarming rate over the past decade, and they are projected to continue to rise in the years to come[2]. Thus early, accurate and efficient diagnosis of skin cancer is warranted. While the current gold standard for diagnosis is histopathological analysis of a biopsy specimen, new imaging technologies such as dermoscopy have made it possible to improve clinical diagnostic accuracy without the need for an invasive biopsy.

Dermoscopy is a diagnostic imaging technique that enables real-time visualization of the microstructures of the epidermis, derma-epidermal junction and papillary dermis at high magnification[3]. In practice, dermoscopy has been shown to improve diagnostic accuracy for several types of cutaneous lesions, most notably skin cancer[4]. However, certain areas may lack the resources to utilize such technology and others do not have access to trained dermatologists. Thus, efforts have been made to utilize artificial intelligence to help diagnose skin lesions from images[6,7]. Most notably, Esteva et al. demonstrated that deep convolutional neural networks could classify images of skin lesions at an accuracy comparable to trained dermatologists[4]. It was also proposed that such a tool could be placed on mobile devices, giving patients access to state-of-the art technology for detecting and diagnosing malignant skin lesions[4].

The International Skin Imaging Collaboration (ISIC) archive contains an image dataset of over 20,000 images of various skin conditions obtained from institutions around the world. Using a subset of these images (HAM10000 Dataset), our goal was to accurately classify dermoscopy images of skin lesions into one of seven categories: melanocytic nevus, dermatofibroma, melanoma, actinic keratosis, basal cell carcinoma, benign keratosis, and vascular lesion. In this manuscript we discuss the design and implementation of our algorithm for this particular task.

## METHODS

**Dataset**

The dataset for the Disease Classification task was downloaded from the ISIC website. The dataset consisted of training (n=10015), validation (n=193), and test (n=1512) image sets obtained from the HAM10000 Dataset[8,9]. Dermoscopy images from all anatomic sites except mucosa and nails were included in this dataset. Each image within the training set had a ground-truth category assignment based on either histopathology, reflectance confocal microscopy, or consensus of at least three expert dermatologists. All malignant cases were confirmed with the gold-standard histopathology.

**Preprocessing of dataset**

All images within the dataset were resized to a 224x224 pixel size. This image size struck a balance between providing a high enough resolution for detection of subtle features on dermoscopy by the model and efficient training. All images were normalized to ImageNet standards. To increase the size of our dataset, we used data augmentation, specifically vertical and horizontal flipping as well as a random zoom up to 1.1x.

**Model Design**

We utilized the Fast.ai framework for designing, training, and evaluating our classifier model for this task[10]. Fast.ai is built on top of PyTorch, which is a widely used deep learning framework for Python[10,11]. We chose Fast.ai specifically for its simplicity and practical use of state-of-the-art techniques such as cyclical learning rates[12,13], differential learning rates, and test-time augmentation. Our algorithm was implemented on Google Colaboratory using an Nvidia Tesla K80 GPU[14]. For our model, we utilized an architecture that made use of deep convolutional neural networks. We employed a strategy known as transfer learning, which is using an existing network trained for a certain task and repurposing it for a similar task. ResNet50[15] is a residual network, which is type of deep convolutional neural network that incorporates residual connections for improved performance. We used the ResNet50 architecture and weights that were previously trained on ImageNet, a database of more than 14 million images of various categories. The final layer was discarded and replaced with a concatenation of an adaptive max pooling layer and an adaptive average pooling layer. This was followed by two fully-connected hidden layers joined by batch normalization layers and finally an output layer consisting of seven units corresponding to the seven disease categories. Dropout layers were used to avoid overfitting of the model to the training dataset. The architecture of our custom model is detailed in Figure 2.

**Training and Evaluation**

Initially, we froze the weights of the ResNet50 architecture within our model and trained the custom top-most layers for 4 epochs or complete iterations through the dataset. Next, we performed fine-tuning – we unfroze the weights and trained the entire model for 15 epochs. A batch size of 32 was used. We used cyclical learning rates with a base learning rate of $1 \times 10^{-2}$

(Figure 3B). Each cycle lasted one epoch, and the learning rate ranged from zero to the base learning rate of $1 \times 10^{-2}$. For the fine-tuning process, we applied differential learning rates across the entire model architecture. The base learning rate was reduced by a factor of 9 for the bottom-third of our model, as these features were more abstract and did not require significant fine-tuning. The learning rate was reduced by a factor of 3 for the middle-third of the model and was kept the same for top third of the model. During the fine-tuning process, each subsequent cycle was prolonged by a factor of 2, as shown in Figure 3C.

After training, we evaluated the model on the validation and test sets. Probabilities of each image belonging to a specific class were calculated. The predicted class for a specific image was the class with the highest probability. In order to further improve the accuracy of our model, we utilized test-time augmentation (TTA). TTA is a technique where the final prediction is based on an average of the prediction made on the original image and the predictions made on random transformations of that image[16].

**RESULTS**

We trained our custom model on the training dataset for a total of 19 epochs (8 cycles). The results were as follows

    Training Loss: **0.148**

    Validation accuracy (no TTA): **88.3%**

    Validation accuracy (TTA): **91.0%**

    Test accuracy (TTA): **73.5%**

TTA resulted in a 2.7% increase in balanced accuracy of the validation set. A plot of the training loss is seen in Figure 3A.

**DISCUSSION**

Our goal was to accurately classify dermoscopy images of skin lesions into one of seven categories. We accomplished just this using a custom model based off of the ResNet50 architecture. Our model was able to correctly diagnose dermoscopy images of skin lesions with a balanced accuracy of 91% on the validation set and 73.5% on the test set after only training for 19 epochs or 8 cycles. Thanks to techniques such as cyclical learning rates, differential learning rates, and TTA, it was possible to enhance the accuracy of model without sacrificing training efficiency. The use of test-time augmentation reduced the misclassification error of our model and improved its overall robustness while cyclical and differential learning rates enabled faster convergence. Given the marked difference in prediction accuracies between the validation and test sets, a significant degree of overfitting exists. Our next steps are to incorporate patient demographic and clinical data, perform additional hyperparameter optimization for further control of overfitting, and explore other architectures to further enhance the performance of our model.

**FIGURES**

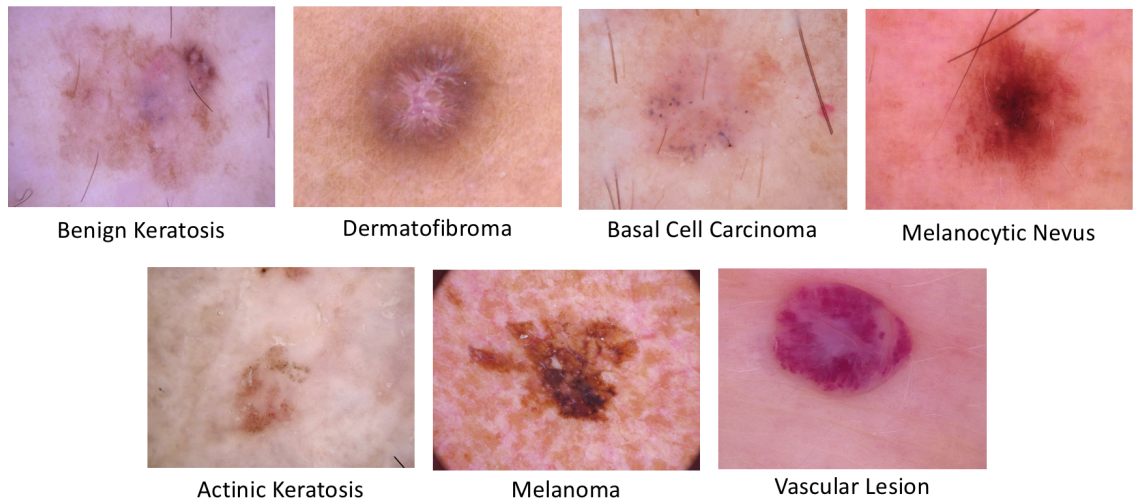

**Figure 1.** Representative image of each class within the HAM10000 dataset.

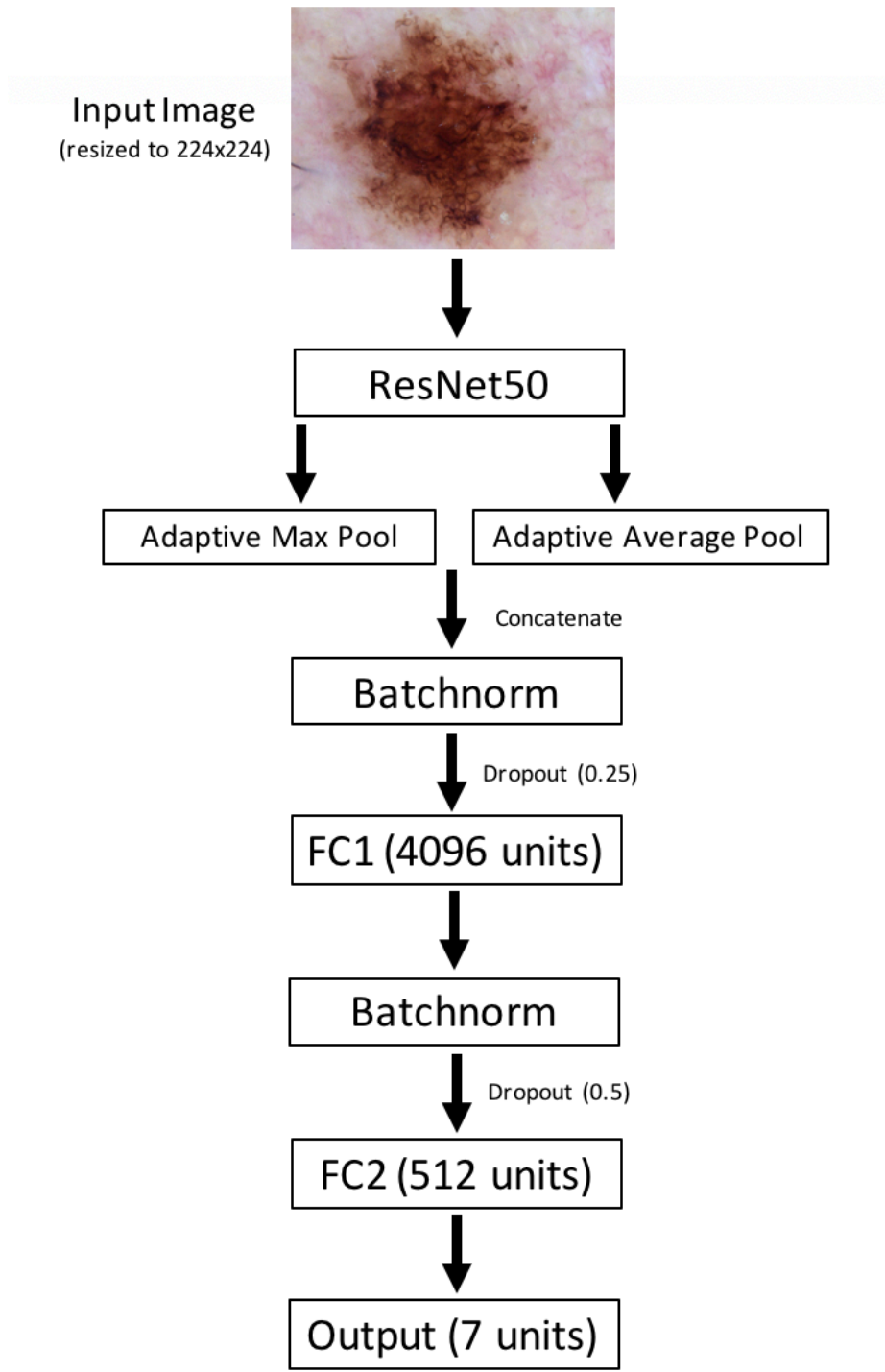

**Figure 2.** Illustration of the architecture of the custom model used for the disease classification task (Task 3).

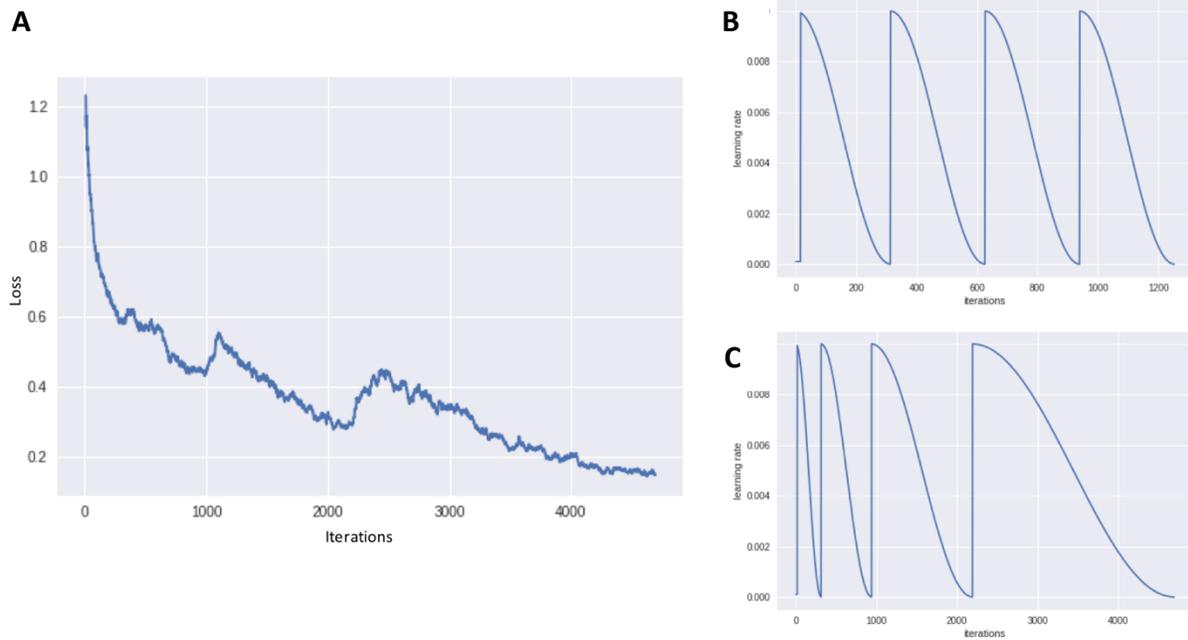

**Figure 3.** A plot of the recorded loss during training process (A). Plots of the learning rate for the initial training of 4 epochs (B) and the fine-tuning process (C).